# A Deep Learning Approach to Identify Rock Bolts in Complex 3D Point Clouds of Underground Mines Captured Using Mobile Laser Scanners

Dibyayan Patra[1], Pasindu Ranasinghe[1], Bikram Banerjee[2], Simit Raval[1,*]

[1]School of Minerals and Energy Resources Engineering, University of New South Wales, Sydney, NSW, Australia.

[2]School of Surveying and Built Environment, University of Southern Queensland, Toowoomba, QLD, Australia.

[*]Corresponding Author Email: simit@unsw.edu.au

**Abstract:** Rock bolts are crucial components of the subterranean support systems in underground mines that provide adequate structural reinforcement to the rock mass to prevent unforeseen hazards like rockfalls. This makes frequent assessments of such bolts critical for maintaining rock mass stability and minimising risks in underground mining operations. Where manual surveying of rock bolts is challenging due to the low light conditions in the underground mines and the time-intensive nature of the process, automated detection of rock bolts serves as a plausible solution. To that end, this study focuses on the automatic identification of rock bolts within medium to large-scale 3D point clouds obtained from underground mines using mobile laser scanners. Existing techniques for automated rock bolt identification primarily rely on feature engineering and traditional machine learning approaches. However, such techniques lack robustness as these point clouds present several challenges due to data noise, varying environments, and complex surrounding structures. Moreover, the target rock bolts are extremely small objects within large-scale point clouds and are often partially obscured due to the application of reinforcement shotcrete. Addressing these challenges, this paper proposes an approach termed *DeepBolt*, which employs a novel two-stage deep learning architecture specifically designed for handling severe class imbalance for the automatic and efficient identification of rock bolts in complex 3D point clouds. The proposed method surpasses state-of-the-art semantic segmentation models by up to 42.5% in Intersection over Union (IoU) for rock bolt points. Additionally, it outperforms existing rock bolt identification techniques, achieving a 96.41% precision and 96.96% recall in classifying rock bolts, demonstrating its robustness and effectiveness in complex underground environments.

## 1. Introduction

In underground mining, movements in the rock strata caused by ground movements, expansions and other factors can lead to hazardous incidents such as rock falls. To mitigate such incidents, rock/cable bolts, shotcrete linings and steel meshes are used to support loose rock mass. Rock bolts are generally steel bars that are bolted into the weak rock strata to provide anchorage and strengthen their load-bearing capacity, hence providing structural support to the walls and roofs of the mine, thereby preventing it from collapsing [1, 2]. Rock falls not only lead to disruptions in mining operations but also can cause fatal injuries and massive damage to mine infrastructure and ventilation systems. Various factors like improper installation, design, and physical changes due to natural as well as mining activities can lead to quality degradation and corrosion in rock bolts [3]. Therefore, it is highly imperative to perform regular assessments of these rock bolts to avert any hazards. Manual investigations of rock bolts are done by surveyors in underground mines, which is extremely time-consuming and challenging due to the low-light conditions in underground mines as well as limited due to restrictive mine access rules. On the other hand, instrumented rock bolts using sensors like strain gauges [4], washer compression sensors [5], load sensors, ultrasonic monitoring [6] and fibre Bragg grating sensors [7] serve as solutions to accurately monitor individual bolts [2, 8]. However, to scale the rock bolt monitoring and visualisation process over vast sections of the mines, automated recognition of rock bolts using close-range remote sensing serves as a potential solution. Traditional surface mapping equipment, such as terrestrial laser scanners (TLS), faces limitations from the absence of GNSS and restricted sensor mobility in underground environments [9]. Sensor mobility is especially needed for scalability in underground mines, as static scanners can only gather data from a fixed vantage point, effectively seeing in one direction. While this may be suitable in open spaces, in underground mines, it results in limited coverage and occlusion of critical regions, restricting the ability to comprehensively map large, complex environments. Additionally, GNSS-based systems are unsuitable in underground environments because satellite signals cannot penetrate dense rock and overburden, preventing the receiver from establishing the satellite communication required for accurate positioning, making the system unscalable across large and complex tunnel networks. With recent advancements in mobile laser scanning and simultaneous localisation and mapping (SLAM), the potential for large-scale 3D point cloud data acquisition in GNSS-denied underground mining sections like tunnels and roadways has been significantly enhanced [10], paving the way for more efficient underground environment mapping and monitoring.

Martínez-Sánchez et al. [11] presented one of the earliest studies incorporating rock bolt identification by registering initial and post-shotcrete point clouds using bolts as reference points. Although rock bolt detection was not their primary goal, their encoder model, comprising an output layer and two autoencoders, targeted bolt identification. To enable accurate shotcrete volume analysis, the method prioritised high recall over precision, leading to many false positives. Despite this trade-off, their work was the first to demonstrate the feasibility of using geometric neighbourhood-based machine learning algorithms for support structure monitoring challenges. In another study, Gallwey et al. [12] explored the use of a machine learning-based approach to identify rock bolts in a point cloud obtained using TLS. They generated a point descriptor comprising of 65 features representing neighbourhood variations like eigenvalue and density-based features, for each point in the point cloud. To classify rock bolts, they tested a neural network and a random forest classifier and found the former to be more accurate. They trained their neural network model on the 65-dimensional classifier descriptor that they manually defined, rather than being learned by a deep neural network. Despite this limitation, they were able to achieve a respectable rock-bolt classification outcome, however, though point-wise performance still left room for improvement.

Using a wide array of features and point descriptors does not necessarily translate to a better classification result since many features might behave similarly for both bolts and non-bolts. For instance, eigenvalue-derived metrics like omnivariance can be prominent around both bolts and rock mass discontinuities, and features such as verticality are ineffective in distinguishing bolts [12]. A different study by Singh et al. [13] used selective point descriptors for rock bolt classification in an MLS-acquired point cloud of an underground mine. Exploiting the cylindrical nature of rock bolts in a point cloud, they used three descriptors, proportion of variance (PoV), radial surface descriptor (RSD) and fast point feature histogram (FPFH), to train their feed-forward artificial neural network (ANN) for rock bolt classification. Later Saydam et al. [14] proposed a bolt detection algorithm tailored for clean small-scale TLS point clouds of construction tunnels prior to shotcrete application, where environmental variability is minimal. Their method followed a coarse-to-fine strategy, initially using the proportion of variance (PoV) descriptor to roughly isolate potential bolt clusters. A neural network was then applied to each cluster, standardised to a fixed size, to distinguish bolt points from the background. This hierarchical approach significantly improved bolt detection precision, thanks to the initial filtering step. In another study, Singh et al. [15] investigated a robust approach for rock bolt identification in MLS point cloud by first introducing a point cloud resampling pre-processing step using moving least squares to mitigate the noisy and sparse nature of an MLS point cloud. Following that, the approach uses a Canupo classifier trained on a multi-scale point descriptor comprising pointness, curveness and surfaceness paired with RANSAC shape detection algorithm. This approach effectively identified bolts by leveraging their cylindrical characteristics and separating them from surrounding mine surfaces.

Most research on rock bolt identification relies on feature engineering and traditional machine learning algorithms. Deep learning methods, on the other hand, offer greater flexibility by automatically learning complex features and typically achieve better performance in irregular geometries [16], making them well-suited for underground mining environments. Although there has been significant progress in deep learning-based semantic segmentation of point clouds [17, 18], limited research has been undertaken to utilise those models in the context of rock bolt identification and support structure mapping. Deep learning models have the inherent ability to identify feature vectors for associating classes and labels with points. Unlike other machine learning algorithms, they don't require manual feature definition. However, specific challenges arise when identifying small-scale objects, such as rock bolts, within large-scale point clouds from underground mines that need to be addressed in order to improve overall accuracy and efficiency.

To this end, this study proposes a novel deep learning architecture, *DeepBolt*, for the automatic identification of rock bolts in medium- to large-scale, complex 3D point clouds of underground mines captured using mobile laser scanners. *DeepBolt* adopts a two-stage approach that combines a geometry-sensitive filtering strategy with a graph-based semantic segmentation model that dynamically constructs local graphs in feature space to capture both local and global geometric structures critical for accurate point-wise classification, as detailed in Section 2. The proposed approach offers a robust and efficient solution to the unique problem of rock bolt identification, which includes severe class imbalance due to the small size of bolts relative to the scan area, data noise and high environmental variability in real-world underground mining scenarios, and visibility of rock bolts due to face-plate obscurity caused by shotcrete application. To assess the performance of the proposed rock bolt identification approach, it is compared to the state-of-the-art semantic segmentation models and current rock bolt identification techniques, as discussed in Section 3.

## 2. Materials and Methods

This section provides a description about the specifications of the 3D point cloud dataset, the technologies and sensors used for their acquisition from underground mines, the pre-processing and noise filtering steps used on the dataset, and the deep learning model used for rock bolt identification in the point cloud.

*2.1. Mobile Laser Scanner*

LiDAR's used in underground mining can be bifurcated into two primary types: TLS (Terrestrial Laser Scanners and MLS (Mobile Laser Scanners) [9, 19-21]. TLS scans the surrounding environment from a fixed point. MLS, on the other hand, is generally hand-held or vehicle-mounted and uses SLAM algorithms (Simultaneous Localisation and Mapping) for scanning. While TLS offers millimetre-level accuracy from a stationary perspective, MLS brings the advantage of mobility, albeit with a compromise in accuracy, which is mostly in centimetres, and increased susceptibility to noise. With increased mobility, MLS has the potential to cut down on blind spots and occlusions, making it a great option for data collection in large-scale regions. Both TLS and MLS have their benefits, and the choice of laser scanner comes down to the application it is being used for.

In terrestrial laser scanning (TLS), the sensor remains stationary at a fixed location, capturing a 3D point cloud of the surrounding environment. In contrast, mobile laser scanning (MLS) involves a moving sensor, where both the sensor position and the environment are initially unknown. Consequently, MLS requires solving two problems: localisation and mapping. Simultaneous Localisation and Mapping (SLAM) addresses these by concurrently estimating the sensor pose and constructing the map SLAM is particularly valuable for rapid mobile scanning in environments with limited or no GNSS availability, such as underground mines, by integrating inertial measurement unit (IMU) data with laser scans to reduce drift and uncertainty [22]. Although SLAM enables efficient underground scanning, it is an optimisation problem that can accumulate position drifts over time, which can be mitigated using loop closure, a process that corrects accumulated drift by recognising previously scanned locations and aligning them accurately.

To collect medium to large-scale data from underground mines, this study uses a handheld mobile laser scanner, Zeb-REVO, that addresses the challenges of immobility and occlusion in TLS data. The system (Table 1) includes:

- A fully integrated Hokuyo UTM-30LX-F scanner fixed on a mechanical 360° rotating head, inertial measurement unit (IMU) and an onboard computer. The scanner is a 2D time-of-flight laser sensor with a rotating head, therefore spinning the sensor around its horizon produces a 3D spherical field of view.

- A SLAM package named GeoSLAM, for post-processing raw data collected by the IMU and laser scanner to generate a high-resolution 3D point cloud. Accurate co-registration of LiDAR frames in GNSS-denied environments is achieved through SLAM, utilizing a precise embedded IMU chip.

**Table 1.** Specifications of mobile laser scanner used in this study.

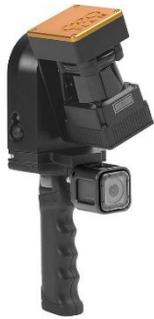

| | Dimensions | 86 × 113 × 287 mm |
|---|---|---|
| | Weight | 1.82 kg |
| | Laser type | 905 nm Class 1 Eye-safe |
| | Laser range | 30 m |
| | Range accuracy | ± 3 cm |
| | Angular resolution | 0.625° horizontal<br>1.8° vertical |
| | Scan rate | 43200 points/s |
| | Power consumption | 30W nominal |
| **Zeb-REVO** | Sensors and accessories | Hokuyo UTM-30LX-F scanner, IMU, data storage, power supply |

*2.2. Data Description*

Training a deep learning model requires large, labelled data. However, no open-source or proprietary data with labelled rock bolts are currently available. To address this, a mobile laser scanner Zeb-REVO was used to collect 3D point clouds from different test areas in Australian underground mines. The selected test areas included medium to large-scale sections of mine roadways (Figure 1a) and tunnels (Figure 1b). These sections range from 45 m to 55 m in length, with cross-sectional dimensions varying between 3.5 m and 5 m in both height and width.

To minimise drift errors in the collected point clouds, data acquisition was performed using a loop trajectory while operating the scanner in handheld mode at walking speed (Figure 1c). Since the scanner relies on SLAM for co-registering successive frames, dynamic objects can introduce false matches. Therefore, data collection was conducted during periods of minimal activity to reduce unwanted mapping drift. The resulting point clouds obtained using the scanner in handheld mode with loop closure have a nominal point spacing of 8 mm and a point density of 15,625 points/m².

The rock bolts used in the mine sites include resin-anchored and grouted types. Data collection was conducted in complex real-world conditions, i.e., after the application of reinforcement shotcrete following excavation (Figure 1d). As a result, only the partially exposed cylindrical protrusion portions of the bolts are visible in the point cloud, while the face plates are not prominently distinguishable. The point clouds were manually annotated, assigning class label 1 to bolt points and class label 0 to non-bolt points, with meticulous visual inspection to ensure accuracy and minimise mislabelling. The annotation was performed after the data pre-processing steps outlined in Section 2.3. The bolt points labelled as 1 are visualised in red for clarity (Figure 1e). Notably, the labelled bolt points correspond only to the clearly exposed cylindrical protrusion sections, while ambiguous areas like the faceplates, which are either minimally visible or entirely obscured by shotcrete, were excluded from rock bolt annotation.

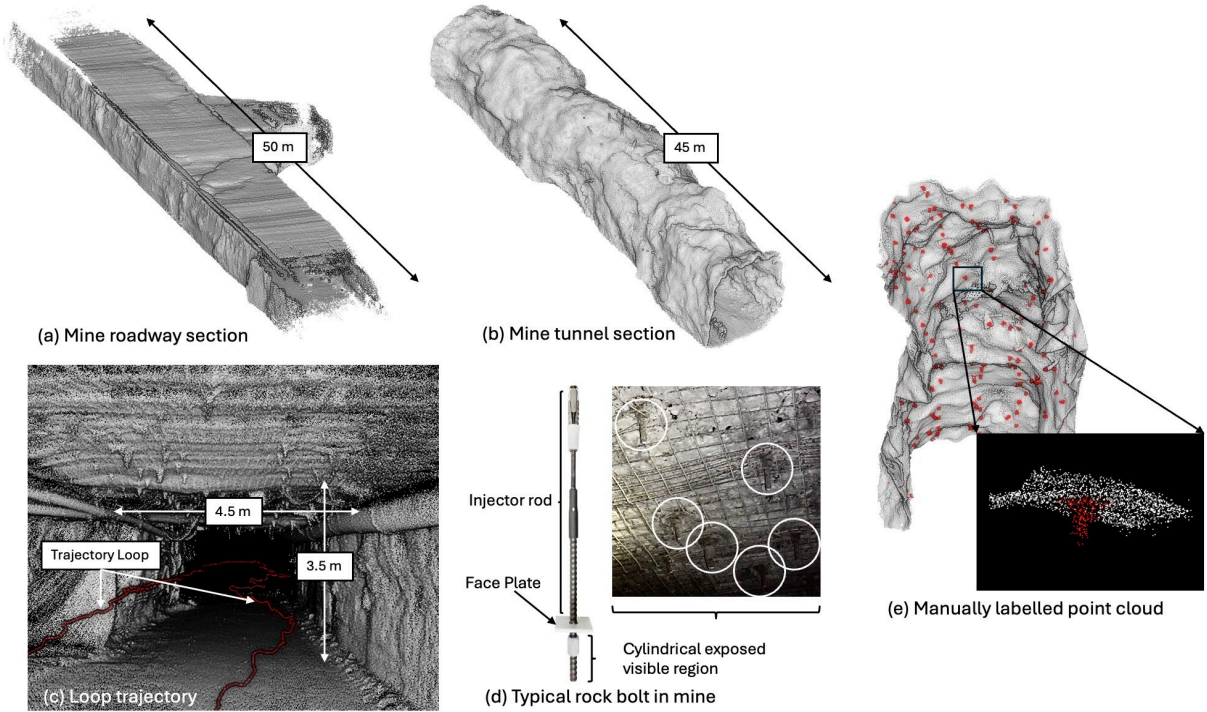

**Figure 1.** Overview of the collected dataset. (a) Sample point cloud of a mine roadway section, (b) Sample point cloud of a mine tunnel section, (c) Cross-section of a sample dataset, illustrating the scanner's loop trajectory in red, (d) A typical rock bolt used in a mine site, including an image highlighting the cylindrical exposed regions visible in real-world conditions; (e) Sample scan visualising the manually annotated rock bolts.

The collected dataset consists of 8 medium to large-scale point cloud scans used in this study. These scans cover diverse sections of mine roadways and tunnels to ensure high generalizability, as the model is exposed to varying rock bolts and a wide range of non-bolt structures. By incorporating scans from distinct regions, the dataset captures real-world variations in bolt positioning, bolt exposure, and surrounding non-bolt structures, ensuring robustness and generalizability of the model by enabling it to learn a wide range of characteristics and diverse scenarios. Each scan contains between 150 and 300 rock bolts, resulting in a total of 1,764 labelled rock bolts across the entire dataset. The surrounding mine environment includes rock surfaces, shotcrete layers, ventilation shafts, cables, markers, and signboards, providing a comprehensive representation of underground mine conditions. The presence of a large number of bolts and highly varying surrounding structures in these medium to large-scale scans enhances the dataset's diversity, making it well-suited for training models capable of accurately identifying rock bolts in complex underground settings. A detailed analysis of the dataset can be seen in Table 2.

**Table 2.** Detailed analysis of the dataset.

| **Total number of rock bolts in the dataset** | 1764 |
|---|---|
| **Average length of scans** | 50 m |
| **Average number of points per scan** | 16M (10M after data pre-processing) |
| **Average number of rock bolt points per scan** | 100k |
| **Average number of rock bolts per scan** | 221 |
| **Average number of points per rock bolt** | 453 |
| **Point spacing** | 8 mm |
| **Point density** | 15,625 points/m$^2$ |

*2.3. Data Pre-processing Steps*

A data pre-processing step is applied to the raw point cloud to eliminate inherent noise and remove unconnected or dynamic objects that appear as spurious points. Additionally, the floor is excluded from the dataset, as rock bolts are typically installed in the walls and roof, not the floor. Including floor data would unnecessarily increase computational load without contributing to the detection process. To carry out this pre-processing, efficient and minimally invasive algorithms such as the k-nearest neighbours (k-NN) filter [23], cloth simulation filter [24], and connected component filter [25] are employed.

The k-nearest neighbours (k-NN) filter is employed in this study to remove isolated erroneous points based on the spatial relationship between each point and its surrounding neighbours. Such noise in the point cloud often arises due to sensor perturbations and laser beam divergence, which can introduce inaccuracies in the range measurements captured by the laser scanner. The algorithm evaluates the local geometry by fitting a plane through the nearest points and identifies outliers based on deviations from this plane. However, selecting an excessively large number of neighbours increases the algorithm's sensitivity to the fitted surface, potentially eliminating valid inlier points and reducing point cloud density. Since plane fitting requires at least four non-collinear points, the number of neighbours must be carefully chosen. Through trial-and-error, it was observed that setting this parameter beyond 15 significantly degraded the dataset. Therefore, the number of neighbours was constrained between 4 and 15, with an optimal value of 6 used in this study.

The Cloth Simulation Filter (CSF) is applied to remove the floor from the scans, which is irrelevant to rock bolt detection. CSF operates by inverting the point cloud and simulating a cloth that is gradually lowered onto the surface under gravity-like forces. As the cloth conforms to the geometry, it separates lower-lying regions from elevated features. The cloth is represented as a grid of interconnected nodes, and its flexibility and resolution are controlled by parameters. In this study, the node spacing of the cloth (grid distance) was set between 25 to 100 times the average point spacing, ensuring an appropriate balance between resolution and computational efficiency. The number of iterations was set between 400 and 500, in line with previous literature suggesting the range sufficient to achieve stable and accurate terrain simulation. Additionally, a threshold distance of 0.5 m from the simulated terrain, empirically derived through trial-and-error, was used to classify and remove floor points, including fragmented rocks and debris commonly found near the base of the scans. This approach effectively segregates the structural surfaces of interest from the ground points.

As a final pre-processing step, connected component filtering is applied to remove isolated sets of points that are disconnected from the main segment of the scanned data. These points often appear as random, isolated sections in the point cloud, resulting from false range measurements, stray objects captured in the scan that are not part of the mine surface, dynamic objects obscured during scanning, or rock debris that passes through the CSF and remains in the floor region. The connected component filtering works by grouping points that are spatially connected based on a defined neighbourhood relationship, typically using an octree structure to efficiently manage spatial data. In this study, an 11th-level octree with a grid size of 0.016 m was used. The choice of an 11th-level octree balances resolution and computational efficiency, maintaining adequate detail without significantly increasing processing time. A minimum threshold of 10,000 points per component was set to ensure that only significant, spatially coherent regions are segmented. Components with fewer points than this threshold are discarded.

These methods are selected over other filtering algorithms as they effectively remove noise and erroneous points while preserving the integrity of the remaining data and requiring relatively low processing time. Importantly, all rock bolts were preserved during pre-processing, and no bolt was unintentionally removed. Only spurious points surrounding the bolt regions, considered as noise, were removed to enhance the definition of the bolts. A comparison between the raw and pre-processed point cloud example scan is illustrated in Figure 2.

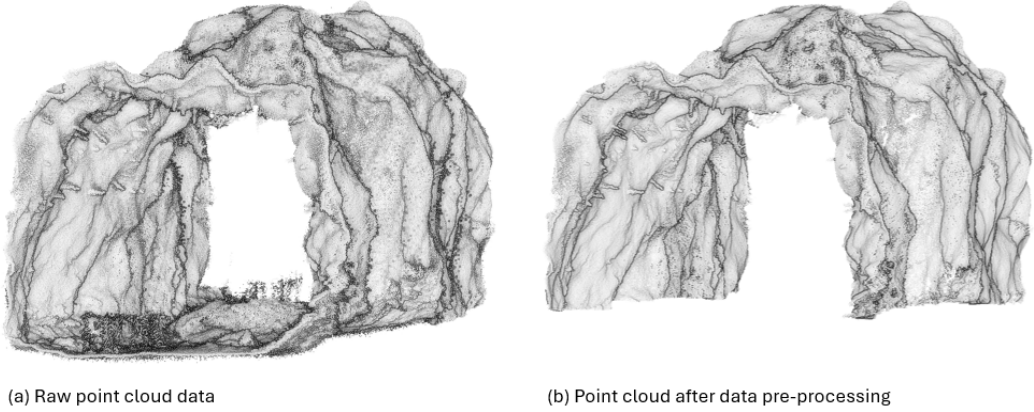

(a) Raw point cloud data          (b) Point cloud after data pre-processing

**Figure 2.** Point cloud data pre-processing showing (a) the raw scan, and (b) the scan after applying pre-processing steps.

*2.4. Deep Learning Model for Rock Bolt Identification*

One of the key advantages of deep learning-based point cloud semantic segmentation models is their ability to automatically learn feature representations for classifying points, eliminating the need for manual feature definition [26, 27]. Traditional machine learning algorithms, in contrast, require predefined features, which is challenging because it is difficult to determine which features are most relevant for distinguishing between classes [28]. Additionally, manually defining features increases processing complexity, as models must operate on a point cloud plus per-point large feature vectors, rather than a simpler point cloud vector, leading to significantly longer processing times.

Despite these advantages, using deep learning to identify rock bolts in complex MLS point clouds of underground mines presents a major challenge due to severe data imbalance. Rock bolts are extremely small objects in large-scale underground point clouds, with non-bolt points (class 0) outnumbering bolt points (class 1) by approximately 100:1 in the pre-processed scans. This imbalance makes it difficult for standard deep learning models to learn meaningful representations for rock bolts. To address these challenges, this paper proposes *DeepBolt*, a novel deep learning architecture designed for automatic rock bolt identification in complex 3D point clouds captured using MLS. *DeepBolt* follows a two-stage approach (Figure 3). In the first stage, it mitigates the data imbalance by filtering out redundant non-bolt points from the environment, increasing the relative proportion of bolt points in the dataset. In the second stage, a graph-based deep learning model, inspired by dynamic graph convolutional neural network [29], performs semantic segmentation to accurately extract rock bolts from the point cloud. This two-stage design enhances both efficiency and accuracy, making *DeepBolt* well-suited for rock bolt detection in challenging underground environments.

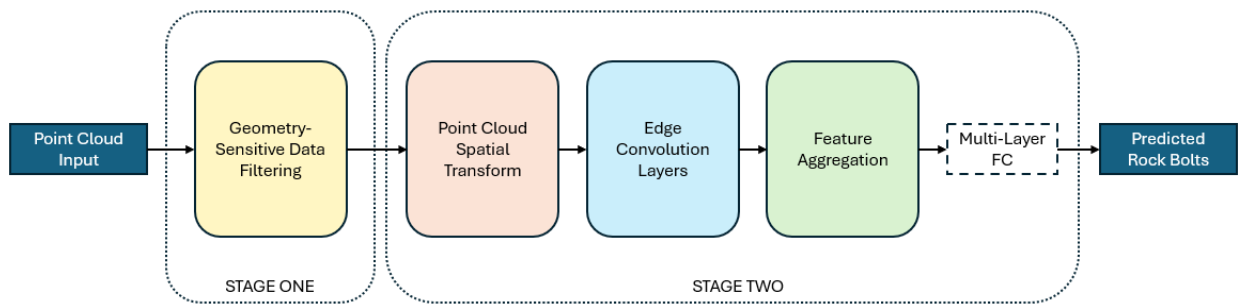

**Figure 3.** Proposed DeepBolt model architecture following a two-stage approach combining geometry-sensitive filtering with graph-based semantic segmentation.

2.4.1. Stage One — Geometry-sensitive Data Filtering

Within the point clouds of underground mines, rock bolts are narrowly perceptible due to their extremely small size, resulting in a significant class imbalance when attempting to identify them. Therefore, a filtering strategy is required to selectively remove unwanted background and non-bolt points without impacting the actual rock bolt points. This would increase the ratio of bolt points to non-bolt points, thereby reducing the data imbalance. To achieve this, a strategy is needed that can roughly differentiate between rock bolts and the surrounding environment. Widely used filtering methods such as uniform voxel downsampling, moving least squares filtering, and region growing filtering are unsuitable for this application, as they cannot distinguish between rock bolts and the environment, leading to similar results across the entire point cloud. In contrast, the geometry-sensitive filtering strategy leverages the cylindrical nature of the visible protrusion sections of rock bolts, which distinguishes them from surrounding elements in the environment. By focusing on this geometric characteristic, the proposed filtering strategy effectively retains rock bolt points while reducing irrelevant background non-bolt points, thereby mitigating the severe class imbalance present in the data.

To understand the geometric characteristics of a surface based on how points are distributed within a local neighbourhood, Principal Component Analysis (PCA) is applied to the point cloud to compute three eigenvalues $\lambda_1$, $\lambda_2$, $\lambda_3$. PCA decomposes the local covariance matrix within a defined support region around each query point, producing three eigenvalues that quantify the variance along the three principal axes [30, 31]. The optimal radius for this support region is strategically determined as $PS.(5 - 16. PS)$, where $PS$ represents the point spacing in meters. This formulation ensures that for lower point spacing, the weight multiplied to the point spacing is larger, while for higher point spacing, it is smaller, adapting dynamically to different scanning resolutions [32]. In this study, with a point spacing of 0.008 m, the computed radius of influence is 0.039 m.

The three eigenvalues characterise the local surface structure, with $\lambda_1$ representing the primary direction of variance, $\lambda_2$ capturing secondary variations, and $\lambda_3$ corresponding to the least significant direction. Using these eigenvalues, key geometric features in geotechnical applications—planarity, omnivariance, and curvature—are derived for each point in the point cloud, as defined in Equation 1. These features provide critical insights into the local shape of a surface.

$$Key\ geometrical\ features \rightarrow \begin{cases} Planarity = \frac{\lambda_2 - \lambda_3}{\lambda_1} \\ Omnivariance = \sqrt[3]{\lambda_1 . \lambda_2 . \lambda_3} \\ Curvature = \frac{\lambda_3}{\lambda_1 + \lambda_2 + \lambda_3} \end{cases} \quad (1)$$

To analyse the effect of the three geometric features on the point cloud, the data is visualised in RGB space, with colours representing the varying degrees of each feature (Figure 4). A detailed analysis of the three features are as follows:

- **Planarity** measures how well a local neighbourhood of points aligns with a plane. As shown in Figure 4a, planarity values are high for well-defined planar surfaces, such as discontinuity planes, and low for non-planar regions, such as edges and bolts. However, while planarity effectively highlights flat surfaces, it is less effective for distinguishing low-planarity regions like rock bolts from other non-planar features, such as discontinuity edges, which also exhibit low planarity.

- **Omnivariance** quantifies the isotropy of the point distribution, capturing how evenly variance is spread across all three principal directions. In Figure 4b, omnivariance values are high for bolts but also remain high in certain discontinuity plane regions, reducing its effectiveness as a distinguishing feature for rock bolts.

- **Curvature** measures the variation along the smallest principal direction, indicating how much a surface deviates from a plane. As seen in Figure 4c, all rock bolts exhibit high curvature values, making curvature the most reliable feature for identifying bolts. Although some other regions such as hanging objects and uneven local surfaces also display moderately high curvature, further refinement in subsequent processing steps, including semantic segmentation, ensures the removal of such unwanted points. Most importantly, all bolts in the point cloud consistently exhibit high curvature, making it a key discriminative feature in the filtering process.

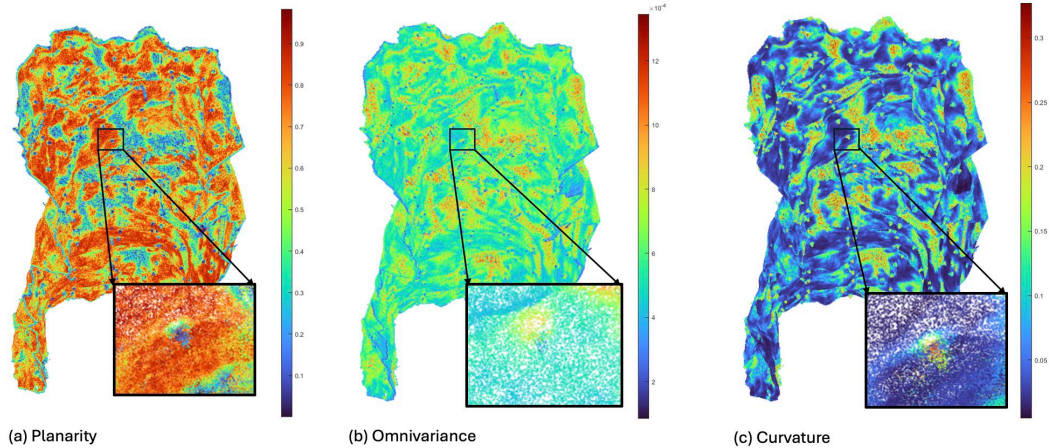

**Figure 4.** Effectiveness of different geometric features in distinguishing rock bolt points. (a) Colorised using RGB based on the degree of Planarity, (b) Colorised using RGB based on the degree of Omnivariance, and (c) Colorised using RGB based on the degree of Curvature.

With curvature identified as the key geometric feature for distinguishing rock bolt regions, an appropriate threshold must be determined to separate regions of interest from background points. This is achieved by computing the 90th percentile of the cumulative distribution function (CDF) of curvature values. The 90th percentile threshold is chosen based on empirical observations, ensuring that 100% of the bolts in the dataset exhibit curvature values above this mark while maintaining a buffer margin for reliability. A fixed curvature threshold is unsuitable due to variations across different datasets. Unlike mean or median based methods, which can be heavily influenced by the dominant background points and lead to overly conservative thresholds that fail to isolate the rare, high-curvature bolt regions, the percentile-based approach specifically targets the upper tail of the distribution where the bolts lie. Instead, this dynamic, percentile-based approach enhances automation while minimising potential errors. The selected 90th percentile value is visualised in the histogram of curvature of one of the scans (Figure 5b), providing an intuitive representation of the threshold selection.

Once established, the threshold effectively separates the point cloud into two categories (Figure 5c):

- **High-curvature points**, consisting of points where curvature exceeds the threshold (Figure 5b). These serve as rough candidates for rock bolts, though they may include some stray high-curvature objects.

- **Low-curvature points**, comprising the remaining points, which predominantly belong to the surrounding environment and are eliminated to improve data quality.

To further refine the selection and filtering process, density-based clustering (DBSCAN) is applied to the set of high-curvature points (Figure 5d). DBSCAN identifies clusters based on spatial proximity, effectively grouping potential rock bolt regions while treating isolated points as noise and removing them. DBSCAN was chosen over other clustering methods because it does not require the number of clusters to be predefined, can identify clusters of arbitrary shape, and is highly robust in segregating noise from clusters in point cloud data. The choice of DBSCAN parameters is guided by key factors such as point density, rock bolt distribution, bolt size, and the average number of points per bolt. Based on these considerations, the maximum neighbourhood distance ε is set to 0.1 m, ensuring that points within a reasonable proximity are grouped together, and the minimum cluster size is set to 50 points, incorporating a buffer for error, as the smallest rock bolts in the dataset typically contain at least 100 points.

To finalise the regions of interest (ROI) as seen in Figure 5e, each high-curvature cluster identified by DBSCAN is evaluated based on two conditions. A rough threshold, $G_{th}$ is set to 400, as no individual rock bolt in the dataset exceeded this size.

- **Clusters larger than $G_{th}$** are directly added to the final filtered point cloud. These clusters are likely to represent localised high-curvature regions that may or may not contain bolts. Retaining them ensures that any bolts within these regions are not inadvertently discarded.

- **Clusters smaller than or equal to $G_{th}$** are refined further. A region of interest (ROI) is defined around the cluster centroid, and all points within a 0.1 m search radius are extracted and added to the filtered point cloud. This step compensates for cases where some portions of a bolt may have been removed from the high curvature set because of exhibiting lower curvature due to local noise, thereby ensuring that entire bolts are preserved. The 0.1 m radius is chosen strategically, as the maximum visible protrusion of a bolt is approximately 0.2 m; thus, a 0.1 m radius around the cluster centroid can effectively capture the full extent of the bolt in all directions while minimising excessive inclusion of extraneous data.

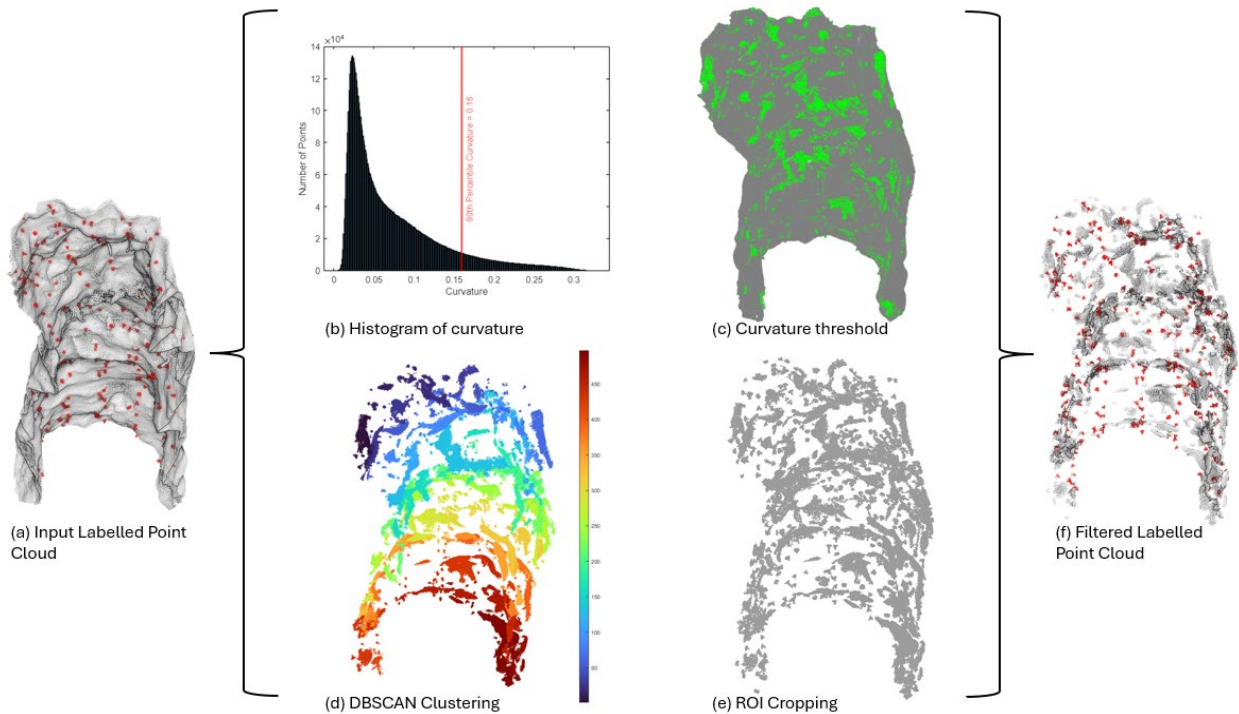

**Figure 5.** Overview of the geometry-sensitive data filtering strategy. (a) Input point cloud with rock bolts marked in red, (b) Histogram of curvature values, with the 90th percentile threshold indicated in red, (c) Curvature thresholding to retain high-curvature regions marked in green, (d) DBSCAN clustering applied to the high-curvature regions for further segmentation and noise removal, (e) Region of Interest (ROI) cropping to remove unwanted background points, (f) Filtered point cloud, preserving rock bolt points while reducing background points.

A comprehensive overview of the geometry-sensitive data filtering strategy is presented in Figure 5. The approach effectively mitigates class imbalance, as evident in the filtered, labelled point cloud (Figure 5f). A substantial portion of background points has been successfully removed while preserving rock bolt points, thereby increasing the ratio of bolt points to non-bolt points. Additionally, the reduction in background points also optimises computational efficiency; fewer points result in faster processing, thereby improving the training and testing times for subsequent semantic segmentation tasks. However, despite the substantial reduction in background points, some amount of non-bolt high-curvature regions persist after the filtering step. These residual points are subsequently addressed through the deep learning-based segmentation model in Section 2.4.2. A detailed pseudocode of the geometry-sensitive data filtering strategy can be seen in Algorithm 1.

---

**Algorithm 1** Geometry Sensitive Data Filtering

    **Input:** point cloud $P$ of an underground mine
    **Output:** filtered point cloud $P_{filtered}$ with reduced class imbalance

1. Compute average point spacing $PS$ using mean distance of points with nearest neighbour
2. Compute radius of influence $r \leftarrow 5*PS - 16*PS^2$
3. **for** each point $p_i \in P$ **do**
    Find neighbouring points of $p_i$ in spherical support region of radius $r$
    Extract and store eigenvalues $\lambda_1, \lambda_2, \lambda_3 \leftarrow$ PCA($p_i$)
    **end for**
4. Compute geometric curvature measure for all points $C \leftarrow \lambda_3 ./ (\lambda_1 + \lambda_2 + \lambda_3)$
5. Compute the cumulative distribution function CDF of $C$
6. Determine the 90th percentile value $C_{th}$ from the CDF
7. Classify points based on $C_{th}$:
    High-curvature points $P_h \leftarrow \{ p_i \mid C_i > C_{th} \}$
    Low-curvature points $P_l \leftarrow P \setminus P_h$
8. Compute clusters $G \leftarrow$ DBSCAN($P_h$)
9. **for** each cluster $g_i \in G$ **do**
    **if** cluster size $|g_i| > G_{th}$ **then**
        $P_{filtered} \leftarrow P_{filtered} + g_i$
    **else**:
        Compute centroid $\mu_i \leftarrow$ mean($g_i$)
        Compute region of interest $P' \leftarrow$ points in $P$ within radius 0.1 to $\mu_i$
        $P_{filtered} \leftarrow P_{filtered} + P'$
    **end if**
    **end for**
10. **return** $P_{filtered}$

---

### 2.4.2. Stage Two — Semantic Segmentation

After applying the filtering process described in Section 2.4.1, the resulting point cloud has significantly reduced background environment (class 0) while enhancing the representation of rock bolts (class 1). This filtering step provides a rough estimation, substantially improving the class imbalance between bolts and non-bolts from approximately 1:100 to 1:15. Achieving this improved balance is critical, as it prevents the segmentation model from being overwhelmed by the dominant non-bolt class, allowing it to better learn and distinguish features associated with rock bolts. This ultimately leads to greater segmentation accuracy, stability, and faster convergence during training.

The next stage involves using semantic segmentation to accurately identify rock bolt protrusions within the filtered point cloud, ultimately isolating the final rock bolts. To achieve this, a graph-based semantic segmentation model inspired by the deep learning model, dynamic graph convolutional neural network (DGCNN), is used in this study. The complete architecture of our model can be seen in Figure 6. Unlike conventional models that rely on CNNs and MLPs, which often struggle to capture the inherent local geometric structures in point clouds [33, 34], our graph-based approach overcomes this limitation by dynamically constructing local neighbourhoods in feature space rather than using fixed receptive fields. By dynamically reconstructing local neighbourhoods in the evolving feature space at each layer, rather than relying on fixed neighbourhoods as in static graph models, the network can better adapt to variations in local geometry and complex structures, and improve feature discrimination between bolts and background. This flexibility enables a more robust feature representation, preserving spatial relationships, adapting to varying point densities, and effectively capturing geometric structures. As a result, the model can accurately distinguish rock bolts from complex backgrounds. The key components of DeepBolt's semantic segmentation model that transform the input into per-point class labels are as follows:

- **Input Layer**: The model takes as input an *n x 6* matrix, where *n* is the total number of points in the filtered point cloud. Each row consists of the spatial coordinates (*X, Y, Z*) and the eigenvalues ($\lambda_1, \lambda_2, \lambda_3$) computed in Section 2.4.1. Incorporating eigenvalues alongside point coordinates enhances the geometric representation of the data, allowing the model to better capture local structural variations. This additional geometric context also accelerates model convergence, as eigenvalues have already been shown to be crucial in characterising the geometric properties of a point within its neighbourhood.

- **Point Cloud Spatial Transform Block**: The point cloud transform block aligns the input point set to a canonical space by estimating and applying a 3 × 3 transformation matrix, ensuring invariance to rotations and translations. To compute this matrix, a tensor is formed by concatenating each point's coordinates with the coordinate differences between the point and its k = 20 nearest neighbours in the k-NN graph. This enriched representation captures both absolute positions and local geometric relationships, allowing the network to learn a transformation that enhances feature consistency for further processing.

- **Edge Convolution Layers (EdgeConv)**: After spatial alignment, the model applies a sequence of three EdgeConv layers, which operate on dynamically computed k-NN graphs (k = 20). Each EdgeConv layer updates the neighbourhood graph at every iteration, enabling the network to capture both local geometric relationships and long-range spatial dependencies. The edge features are processed through a multi-layer perceptron (MLP), which learns non-linear transformations to extract high-level representations from the local neighbourhood. By stacking multiple EdgeConv layers with MLPs, the network progressively refines feature representations.

- **Feature Aggregation**: As the network progresses, multiple EdgeConv layers extract hierarchical features at different levels. Early layers focus on local details, while deeper layers capture global contextual information. The extracted features from different layers are concatenated to preserve both fine and coarse details, ensuring a robust segmentation of rock bolts even in complex underground environments.

- **Fully Connected Layers**: Finally, the model applies fully connected layers to predict per-point class labels. A combination of local and global features is processed through an MLP, which outputs class logits for each point. Instead of softmax, a sigmoid activation function is used, as it produces a single probability for the positive class (rock bolt). The choice of sigmoid over softmax is driven by the binary nature of the classification task, where the model only needs to distinguish between two classes: rock bolt and non-bolt. This simplifies the model, reducing complexity by requiring just one output neuron instead of a neuron for each class, as would be the case with softmax. Sigmoid activation is computationally more efficient, requiring only a single neuron for binary class, and facilitates handling class imbalance through weighted loss functions. Additionally, it avoids unnecessary competition between outputs, resulting in more stable gradients. This activation function enables the final binary segmentation output. By leveraging graph-based dynamic feature extraction, the network ensures accurate segmentation of rock bolts, even in noisy, high-clutter environments.

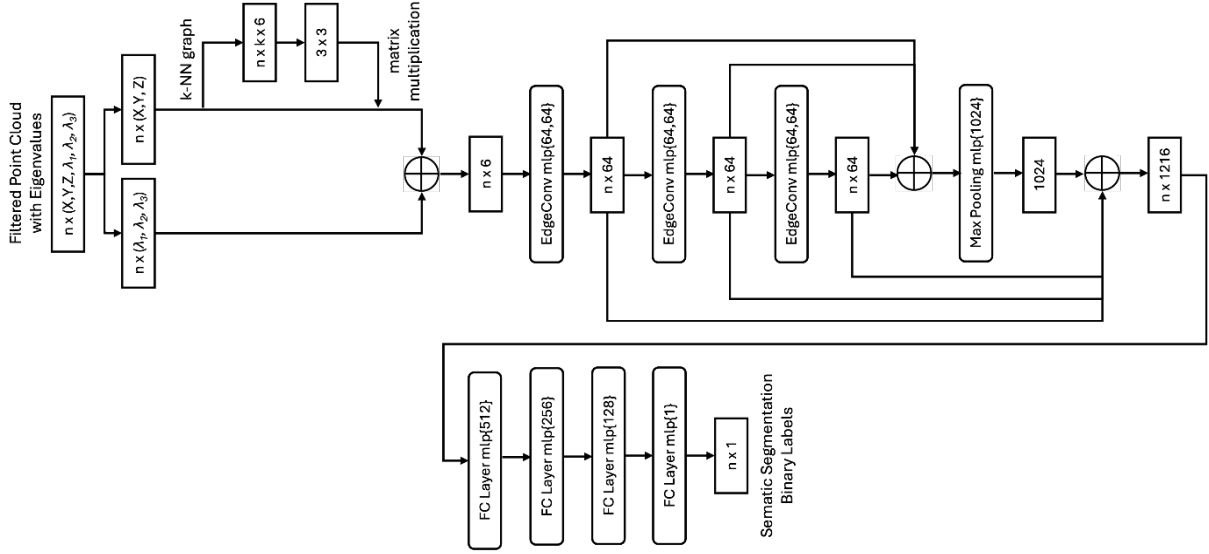

**Figure 6.** Overview of the semantic segmentation architecture used in DeepBolt.

To effectively train the model and refine its segmentation accuracy, the model employs a weighted sigmoid binary cross-entropy loss function. This approach outperforms other loss functions in scenarios involving class imbalance, such as the detection of rock bolts in point cloud data, by addressing the unequal distribution of the target classes. Given that rock bolt points are outnumbered by background points in the input to the model at a ratio of approximately 1:15, the loss function is adjusted to assign a higher penalty to misclassifications of the minority rock bolt class. This weighting ensures that the model learns to correctly identify rock bolts despite their lower representation in the dataset. Accordingly, the loss function applies weights of 15/16 to rock bolt points and 1/16 to background points using a normalised inverse frequency scheme. The mathematical formulation of the loss function used in the model is presented in Equation 2.

$$L = -w^+ y \log(\sigma(x)) - w^- (1-y) \log(1 - \sigma(x)) \qquad (2)$$

where:
- $\sigma(x) = \frac{1}{1+e^{-x}}$ (sigmoid function)
- $y \in \{0,1\}$ (true label)
- x is logits before sigmoid activation
- $w^+$ = 15/16 (weight for rock bolt class)
- $w^-$ = 1/16 (weight for non-bolt class)

As outlined in Section 2.2, the dataset comprises a total of 1,764 rock bolts. To prevent overfitting, the dataset was split into two parts such that filtered point cloud scans consisting of 1,468 bolts were used for training, while the remaining scan with 296 bolts was reserved for testing. A 10-fold cross-validation approach was employed during the training process, where each fold consists of nine subsets used for training and one unique subset for validation. To monitor the stability and convergence of the training, both training and validation loss curves were tracked over 100 epochs. As shown in the learning curve (Figure 7), the training loss exhibits a consistent decline, indicating effective learning, with early training phases showing higher fluctuations due to initial weight updates, but as training progresses, the loss stabilises, signalling convergence. Similarly, the validation loss follows a comparable trend, stabilising similarly to the training loss after several epochs. This parallel behaviour suggests that the model generalises well to unseen data, effectively learning the underlying patterns rather than memorising the training set. As a result, the model does not overfit, and the validation loss decreases steadily, reinforcing the model's ability to predict unseen data. The model's loss plateaus and converges after 32 epochs, establishing this as the optimal number of epochs for training. For optimising the loss function, the Adam optimiser was used with an initial learning rate of 0.001, momentum of 0.1, and a batch size of 16, with momentum decay set to 0.5 every 16 steps. The model was implemented using PyTorch and takes approximately 8 hours to converge in 32 epochs on an NVIDIA T1000 GPU (8GB VRAM, 896 CUDA cores).

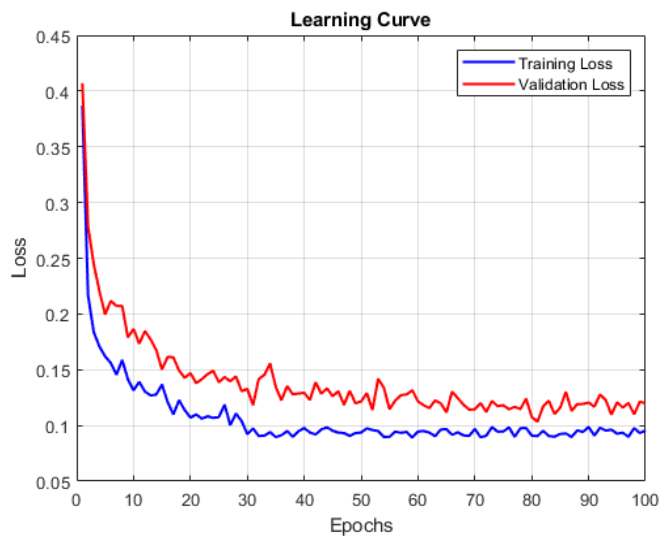

**Figure 7**. Learning curve of the model.

## 3. Results and Discussions

This section provides a comprehensive analysis of the results obtained using the proposed method described in Section 2. The experiments were conducted on a system with the following computational specifications: a 3.90 GHz Intel® Xeon® W-2245 64-bit processor, 128 GB of system memory, and an NVIDIA T1000 GPU with 8 GB VRAM and 896 CUDA cores.

*3.1. Evaluation of Geometry-based Filtering Strategy*

As established in Section 2.4, *DeepBolt* employs a geometry-sensitive filtering strategy to address the massive class imbalance present in the dataset. The effectiveness of this strategy was assessed using three key metrics: (i) the percentage of background points removed, (ii) the percentage of individual rock bolts preserved, and (iii) the average percentage of rock bolt points retained after filtering.

An example scan processed with the filtering strategy is illustrated in Figure 8, highlighting key regions of interest. The visualisations include (a) individual bolt within the region of interest, (b) bolt located in a local high-curvature area, and (c) high-curvature structure that is not a rock bolt but still passes through the filtering process. While large portions of background noise are removed, certain stray high-curvature objects persist. This is because the filtering is based on geometric features, which may not always uniquely identify all background elements from objects of interest. Additionally, the complexity of the rock mass environment, with varying shapes and curvatures, poses challenges in setting optimal thresholds that balance background removal with the retention of key features. These persistent stray background points are subsequently addressed by the semantic segmentation model. The primary goal of the filtering strategy is to reduce the class imbalance while ensuring that no critical bolt information is lost, thus limiting the degree of background removal. These visual results confirm the strategy's ability to retain essential elements while minimising irrelevant background environment.

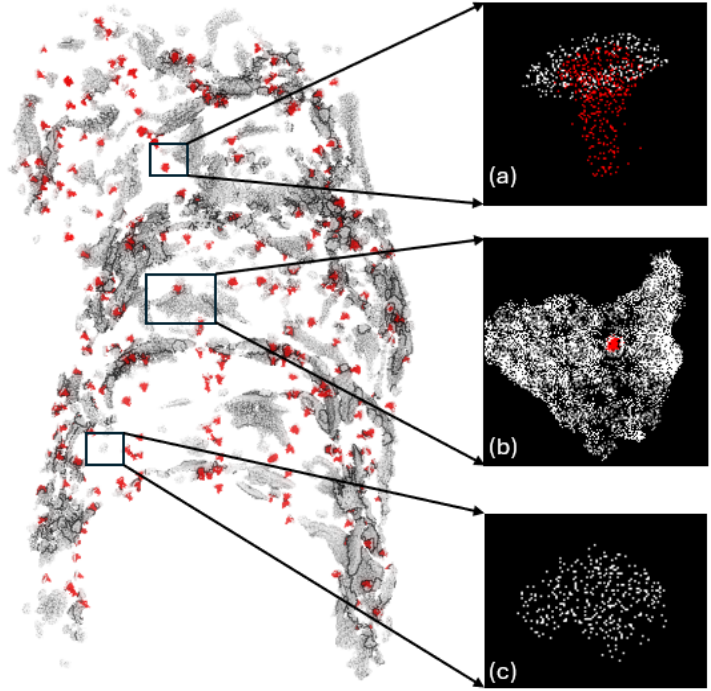

**Figure 8.** Example scan filtered using geometry-sensitive filtering strategy, showing zoomed-in section of (a) individual bolt in ROI, (b) bolt within local high curvature region, and (c) non-bolt high curvature structure.

A quantitative evaluation of the filtering strategy is presented in Table 3, summarising results from eight different scans. On average, 85% of background points were removed, while 98.37% of rock bolt points were preserved. The minor loss in bolt points can be attributed to edge cases where portions of bolts were excluded due to local high-curvature considerations or data noise. However, this minimal loss has a negligible impact on the final segmentation and intersection-over-union results since the overall loss is less than 2%. More importantly, 100% of individual rock bolts were retained in all cases, ensuring that no complete bolts were lost during the filtering process.

The variations observed in the percentage of background removal and bolt point preservation across different scans can be attributed to local scan factors such as variations in curvature, environmental structures, and data noise. Despite these variations, the method consistently achieves over 80% reduction in non-bolt points while preserving over 98% of bolt points, making it a robust and reliable approach for improving the representation of rock bolts in underground mine point clouds. As a result, the class imbalance between bolts and non-bolts is successfully reduced from approximately 1:100 to 1:15, significantly mitigating the severe class imbalance in the dataset for subsequent processing.

**Table 3.** Evaluation of the filtering strategy: percentage of background removed and bolts preserved.

| Dataset | # points in pre-processed scan | # points in filtered scan | % of background removed | % of rock bolts preserved | % of bolt points preserved |
|---|---|---|---|---|---|
| Scan 1 | 8M | 1.3M | 84% | 100% | 98.85% |
| Scan 2 | 9.3M | 1.7M | 82% | 100% | 98.96% |
| Scan 3 | 10.8M | 1.2M | 88% | 100% | 98.39% |
| Scan 4 | 11M | 1.9M | 83% | 100% | 97.72% |
| Scan 5 | 7.7M | 1.3M | 83% | 100% | 98.73% |
| Scan 6 | 13.7M | 2M | 85% | 100% | 98.59% |

| Scan 7 | 7.3M | 0.8M | 89% | 100% | 97.81% |
| Scan 8 | 12.2M | 1.7M | 86% | 100% | 97.90% |
| **Average** | **~10M** | **~1.5M** | **~85%** | **100%** | **98.37%** |

*3.2. Evaluation of Semantic Segmentation and Overall Rock Bolt Identification Performance*

To evaluate the semantic segmentation performance of *DeepBolt*, it is compared against several state-of-the-art, widely adopted deep learning models, as outlined below.

- PointNet [35]: A deep learning architecture for point cloud classification and segmentation that directly processes raw 3D points, capturing global geometric features.

- PointNet++ [36]: An extension of PointNet that captures local context by hierarchically aggregating features through a series of spatially grouped regions, improving performance on complex structures.

- KPConv [37]: A kernel-based convolution approach for 3D point clouds that uses learnable convolution kernels, allowing for more flexible and efficient feature extraction on irregular point sets.

- RandLA-Net [38]: A lightweight, efficient deep learning model for large-scale 3D point cloud segmentation, utilizing random sampling and local aggregation to achieve high accuracy with reduced computational cost.

- DGCNN [29]: A dynamic graph convolutional network that models point cloud data through dynamic graph structures, enabling robust feature extraction by considering the relationships between points.

The semantic segmentation performance of these models is evaluated using the Intersection over Union (IoU) metric for both bolts and non-bolts. IoU measures the overlap between the predicted segmentation and the ground truth, indicating how well the model identifies each class. The formula for calculating IoU for a given class (bolt or non-bolt) is provided in Equation 3. The rock bolt semantic segmentation results of *DeepBolt*, compared to other models, are shown in Figure 9. The observed performance differences between the compared models stem from variations in local feature aggregation, robustness to class imbalance, and adaptability to the complex spatial distributions typical of underground mine point clouds. *DeepBolt* outperforms all other models in terms of both bolt and non-bolt IoU's. Specifically, it achieves a significant improvement in bolt IoU, with an increase ranging from 29.5% to 42.5%. This enhancement is attributed to the fact that existing models are not optimised to handle severe class imbalance or detect small objects like rock bolts in large-scale, complex point clouds of underground mines. Unsurprisingly, the gain in non-bolt IoU is relatively modest, approximately 0.4%, since over 99% of the point cloud consists of background environment. Overall, the IoU results highlight *DeepBolt's* superior capability in accurately segmenting rock bolts in complex environments.

$$IoU = \frac{|P \cap G|}{|P \cup G|} \qquad (3)$$

where:
- *P* is the predicted segments of a given class
- *G* is the ground truth segments of a given class

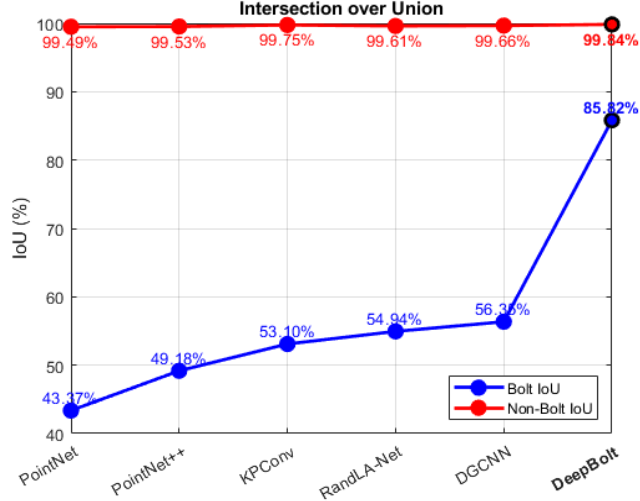

**Figure 9.** Semantic segmentation performance evaluation using Intersection over Union (IoU) metric.

The semantic segmentation and the overall rock bolt identification performance of DeepBolt can be visualised in (Figure 10). The ground truth rock bolts and the predicted rock bolts are marked in red and green respectively in the testing data. The figure includes an example where a predicted bolt is overlaid on the ground truth, clearly demonstrating the accuracy of the segmentation. The predicted bolt aligns closely with the ground truth, with only minor discrepancies observed near the base of the bolt. These small differences likely stem from slight variations in the manual labelling of the point cloud and minor inaccuracies in prediction. Importantly, there is no noticeable deviation in the prominent cylindrical section of the bolt. This visual comparison highlights DeepBolt's strong performance and its ability to accurately identify rock bolts in complex 3D environments.

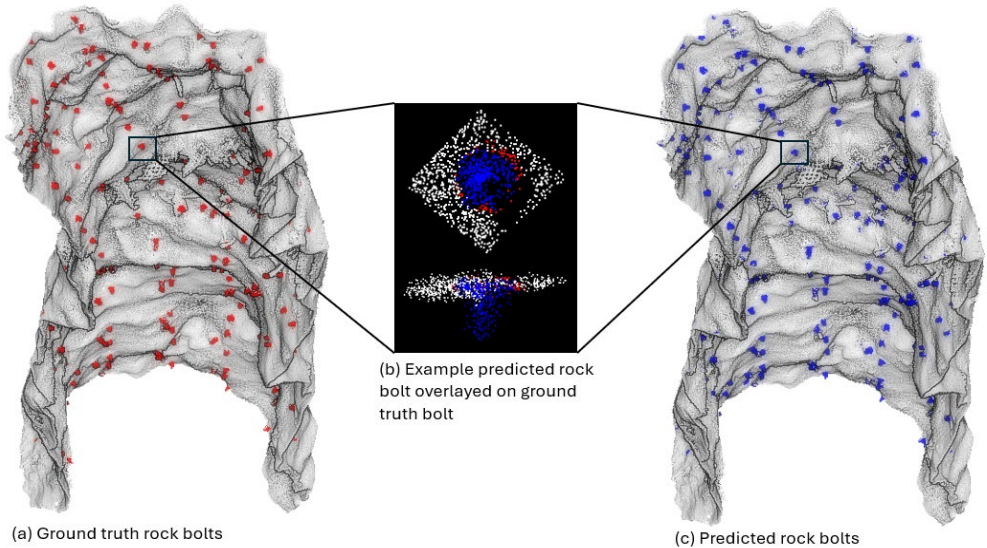

**Figure 10.** Rock bolt identification results in test data. (a) Shows the ground truth rock bolts marked in red. (b) Overlays an example predicted rock bolt on the ground truth to visualise the performance of DeepBolt. (c) Shows the predicted bolts identified by DeepBolt marked in blue.

To assess the overall performance of the proposed *DeepBolt* as a rock bolt identification technique, it is compared against two of the latest techniques in the literature, BoltANN [13] and CanupoBolt [15], as described in Section 1.1. The evaluation is based on standard classification metrics, precision, recall, and F1 score, as defined in Equation 4, using the test dataset containing 296 ground truth bolts. True Positives (TP) refer to bolts correctly identified as such, False Positives (FP) are incorrectly identified bolts that do not correspond to ground truth, and False Negatives (FN) are actual bolts that were missed by the model. Precision measures the proportion of predicted bolts that are actually correct, recall measures the proportion of actual bolts that were successfully identified, and F1 score provides a balanced measure that considers both precision and recall.

$$Classification\ metrics \rightarrow \begin{cases} Precision = \dfrac{TP}{TP+FP} \\ Recall = \dfrac{TP}{TP+FN} \\ F1\ Score = 2 \cdot \dfrac{precision \cdot recall}{precision + recall} \end{cases} \quad (4)$$

A detailed analysis of the rock bolt classification performance can be seen in Table 4. *DeepBolt* achieved the highest number of true positives while minimizing both false positives and false negatives. It delivered an impressive precision of 94.41%, a recall of 96.96%, and an F1 score of 0.96, outperforming both BoltANN and CanupoBolt. While BoltANN and CanupoBolt showed comparable recall values, BoltANN had a slight edge in precision between the two. When compared to its counterparts, *DeepBolt* improved upon BoltANN's precision by 4.88% and CanupoBolt's by 7.57%. In terms of recall, it surpassed BoltANN by 7.43% and CanupoBolt by 7.76%, while the F1 score showed an improvement of 0.06 over BoltANN and 0.08 over CanupoBolt. These results clearly demonstrate *DeepBolt's* superiority in accurately identifying rock bolts within complex 3D point clouds of underground mines, thereby validating the effectiveness of this deep learning-based approach. The lower false positives and false negatives in DeepBolt significantly enhance the practical usability of *DeepBolt* compared to BoltANN and CanupoBolt, ensuring fewer background misclassifications (higher precision) and fewer missed bolts (higher recall), both critical for reliable and real-world application of automated bolt identification.

**Table 4.** Rock bolt classification performance evaluation.

| Method | Ground Truth | True Positive | False Positive | False Negative | Precision % | Recall % | F1 Score |
|---|---|---|---|---|---|---|---|
| BoltANN | 296 | 265 | 31 | 31 | 89.53% | 89.53% | 0.90 |
| CanupoBolt | 296 | 264 | 40 | 32 | 86.84% | 89.20% | 0.88 |
| ***DeepBolt*** | 296 | 287 | 17 | 9 | **94.41%** | **96.96%** | **0.96** |

*3.3. Execution Times*

The average computation time for identifying rock bolts in a test scan using DeepBolt is summarised in Table 5. The table outlines the execution times for each step in the pipeline, excluding the time required to train the semantic segmentation model. Model training takes approximately 8 hours to converge over 32 epochs. While the geometry-sensitive filtering strategy incurs a relatively high computational cost, it plays a critical role in addressing the severe class imbalance by significantly reducing background points. This reduction improves the efficiency and accuracy of the subsequent semantic segmentation step. The entire process of identifying rock bolts in a test scan covering ~45m of real-world underground mine section takes approximately 1650 seconds in total. The execution times are limited by the computational resources used for this study and can be further optimised by using upgraded overall hardware, including faster processors, GPUs with more cores and higher memory bandwidth, and improved storage systems to accelerate data handling and parallel processing.

**Table 5.** Stepwise execution times for identifying rock bolts in test scan.

| Process | Time taken (s) |
|---|---|
| Pre-processing – k-NN filtering | 9.27 |
| Pre-processing – Connected component filtering | <1.00 |
| Pre-processing – Cloth simulation filter | <1.00 |
| Geometry-sensitive filtering – Eigenvalue calculations | 682.36 |
| Geometry-sensitive filtering – DBSCAN | 768.89 |
| Geometry-sensitive filtering – ROI cropping | 31.26 |
| Semantic segmentation | 157.41 |
| **Total time taken to identify rock bolts in test scan** | **≈ 1650 seconds** |

*3.4. Rock Bolt Distance and Distribution Maps*

Once rock bolts are accurately identified in the point cloud of underground mine sections, their spatial locations can be utilised to analyse the overall distribution and density of rock bolts across the scanned area. Using the identified bolt positions, two key visualisations are generated: the rock bolt distance map and the rock bolt distribution map, as shown in Figure 11. The distance map (Figure 11a) reflects how far each point in the point cloud is from the nearest bolt. Blue regions indicate close proximity (less than 0.6 m), while red areas represent points farther than 1.4 m from any identified bolt. Complementing this, the distribution map (Figure 11b) conveys the density of bolts around each point by indicating how many bolts are found within a unit radius (set to 2 meters in this study). In this map, blue denotes areas with high bolt concentration (more than 16 bolts nearby), and red signifies sparse coverage (fewer than 6 bolts).

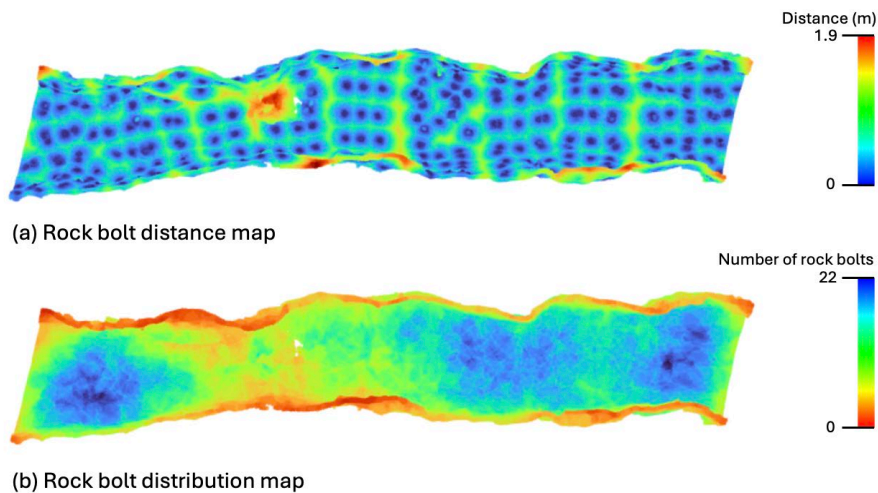

**Figure 11.** (a) Rock bolt distance, and (b) rock distribution maps generated from identified bolt locations.

The accurate and automatic identification of rock bolts enables precise localisation of bolt positions within point clouds of underground mine sections. This capability forms the foundation for generating rock bolt distance and distribution maps, which together provide a comprehensive visualisation of the installed rock bolt support system. These maps are invaluable in informing key mining decisions: they help engineers and mine operators assess whether the existing bolt layout aligns with the intended support design, ensuring that support systems are in place where needed. Distance and distribution maps specifically assist in identifying areas requiring additional reinforcement, enabling efficient re-bolting operations, and ensuring that maintenance efforts are targeted where bolt presence is insufficient or deteriorated. By comparing actual bolt placements with mine support installation plans, these maps can reveal deviations from the original design, pinpointing areas of under-support or potential failures. Moreover, distribution and density maps can highlight changes in bolt presence over time, such as regions where bolts may have been dislodged, degraded, or lost, particularly in older mine sections. This capability is crucial for proactive maintenance and for identifying risk areas before they lead to safety concerns. With an automated approach like DeepBolt, using mobile laser scanning, such monitoring becomes not only more efficient and scalable but also significantly less labour-intensive compared to traditional manual inspections. As a result, the rock bolt identification technique offered by DeepBolt provides a practical, time-saving solution that greatly enhances the safety and integrity of underground mine environments by ensuring continuous and accurate monitoring of the rock bolt support system.

## 4. Conclusions

Rock bolts play a vital role in ensuring the structural integrity of underground mines by reinforcing weak rock strata and preventing hazardous events such as rockfalls. However, the challenging nature of manual rock bolt assessment necessitates an automated and scalable alternative. This paper presented a novel deep learning approach *DeepBolt,* for the automatic identification of rock bolts in complex, medium- to large-scale 3D point clouds captured using SLAM based mobile laser scanners in real-world underground mining environments. DeepBolt addresses several critical challenges in this domain, including severe class imbalance, high environmental variability, and the partial visibility of rock bolts due to face-plate obscurity caused by shotcrete application. It employs a two-stage approach combining geometry-sensitive filtering with a graph-based semantic segmentation network, specifically optimised for detecting small rock bolt objects in large-scale point clouds. Evaluation results demonstrated that our proposed architecture outperforms both state-of-the-art deep learning semantic segmentation models and current rock bolt identification techniques. The proposed method offers a robust, efficient, and scalable solution for automated rock bolt identification, enabling more frequent and reliable assessments of mine support systems. Future work can extend the *DeepBolt* training across different mine settings, geological conditions and complex rock formations for further broadening the system's generalisability and applicability. Additionally, this framework can also be extended to real-time bolt monitoring, predictive analytics of bolt conditions, multi-temporal analysis, or integration with digital twin environments to further enhance underground mine safety and operational efficiency.

## References

[1] C. C. Li. Principles of rockbolting design. Journal of Rock Mechanics and Geotechnical Engineering 2017, 9, 396-414, doi:10.1016/j.jrmge.2017.04.002.

[2] G. Song, W. Li, B. Wang and S. C. Ho. A Review of Rock Bolt Monitoring Using Smart Sensors. Sensors (Basel) 2017, 17, 776, doi:10.3390/s17040776.

[3] N. Aziz, P. Craig, J. Nemcik and F. Hai. Rock bolt corrosion – an experimental study. Mining Technology 2014, 123, 69-77, doi:10.1179/1743286314y.0000000060.

[4] H. Nourizadeh, A. Mirzaghorbanali, N. Aziz, K. McDougall and A. A. Sahebi. Development of a wireless system to measure the strain/deformation of rock bolts. In Proceedings of the Proceedings of the 2022 Resource Operators Conference, University of Wollongong, 2022; pp.

[5] B. Wang, L. Huo, D. Chen, W. Li and G. Song. Impedance-Based Pre-Stress Monitoring of Rock Bolts Using a Piezoceramic-Based Smart Washer-A Feasibility Study. Sensors (Basel) 2017, 17, doi:10.3390/s17020250.

[6] Z. Sun, K.-T. Wu, S. Kruger, D. Levesque, D. Gagnon, Y. Quenneville, R. Lacroix and R. Royer. A new paradigm in ground support monitoring through ultrasonic monitoring of clusters of rockbolts. In Proceedings of the Proceedings of the Ninth International Symposium on Ground Support in Mining and Underground Construction, 2019; pp. 75-84.

[7] Y. Zhao, N. Zhang and G. Si. A Fiber Bragg Grating-Based Monitoring System for Roof Safety Control in Underground Coal Mining. Sensors (Basel) 2016, 16, doi:10.3390/s16101759.

[8] B. Lama and M. Momayez. Review of Non-Destructive Methods for Rock Bolts Condition Evaluation. Mining 2023, 3, 106-120.

[9] S. Raval, B. P. Banerjee, S. Kumar Singh and I. Canbulat. A Preliminary Investigation of Mobile Mapping Technology for Underground Mining. In Proceedings of the IGARSS 2019 - 2019 IEEE International Geoscience and Remote Sensing Symposium, 2019; pp. 6071-6074.

[10] S. K. Singh, B. P. Banerjee and S. Raval. A review of laser scanning for geological and geotechnical applications in underground mining. International Journal of Mining Science and Technology 2023, 33, 133-154, doi:10.1016/j.ijmst.2022.09.022.

[11] J. Martínez-Sánchez, I. Puente, H. GonzálezJorge, B. Riveiro and P. Arias. Automatic Thickness and Volume Estimation of Sprayed Concrete on Anchored Retaining Walls from Terrestrial Lidar Data. The International Archives of the

Photogrammetry, Remote Sensing and Spatial Information Sciences 2016, XLI-B5, 521-526, doi:10.5194/isprs-archives-XLI-B5-521-2016.

[12] J. Gallwey, M. Eyre and J. Coggan. A machine learning approach for the detection of supporting rock bolts from laser scan data in an underground mine. Tunnelling and Underground Space Technology 2021, 107, 103656, doi:10.1016/j.tust.2020.103656.

[13] S. K. Singh, S. Raval and B. Banerjee. Roof bolt identification in underground coal mines from 3D point cloud data using local point descriptors and artificial neural network. International Journal of Remote Sensing 2020, 42, 367-377, doi:10.1080/2150704x.2020.1809734.

[14] S. Saydam, B. Liu, B. Li, W. Zhang, S. K. Singh and S. Raval. A Coarse-to-Fine Approach for Rock Bolt Detection From 3D Point Clouds. IEEE Access 2021, 9, 148873-148883, doi:10.1109/access.2021.3120207.

[15] S. K. Singh, S. Raval and B. Banerjee. A robust approach to identify roof bolts in 3D point cloud data captured from a mobile laser scanner. International Journal of Mining Science and Technology 2021, 31, 303-312, doi:10.1016/j.ijmst.2021.01.001.

[16] Y. Guo, H. Wang, Q. Hu, H. Liu, L. Liu and M. Bennamoun. Deep Learning for 3D Point Clouds: A Survey. IEEE Trans Pattern Anal Mach Intell 2021, 43, 4338-4364, doi:10.1109/TPAMI.2020.3005434.

[17] Y. Xie, J. Tian and X. X. Zhu. Linking Points With Labels in 3D: A Review of Point Cloud Semantic Segmentation. IEEE Geoscience and Remote Sensing Magazine 2020, 8, 38-59, doi:10.1109/mgrs.2019.2937630.

[18] R. Zhang, Y. Wu, W. Jin and X. Meng. Deep-Learning-Based Point Cloud Semantic Segmentation: A Survey. Electronics 2023, 12, doi:10.3390/electronics12173642.

[19] U. Wong, A. Morris, C. Lea, J. Lee, C. Whittaker, B. Garney and R. Whittaker. Comparative evaluation of range sensing technologies for underground void modeling. In Proceedings of the 2011 IEEE/RSJ International Conference on Intelligent Robots and Systems, 2011; pp. 3816-3823.

[20] B. KekeÇ, N. BİLİM, E. Karakaya and D. GhİLoufİ. Applications of Terrestrial Laser Scanning (Tls) in Mining: A Review. Turkey Lidar Journal 2021, doi:10.51946/melid.927270.

[21] W. Mukupa, G. W. Roberts, C. M. Hancock and K. Al-Manasir. A review of the use of terrestrial laser scanning application for change detection and deformation monitoring of structures. Survey Review 2016, 1-18, doi:10.1080/00396265.2015.1133039.

[22] C. Cadena, L. Carlone, H. Carrillo, Y. Latif, D. Scaramuzza, J. Neira, I. Reid and J. J. Leonard. Past, Present, and Future of Simultaneous Localization and Mapping: Toward the Robust-Perception Age. IEEE Transactions on Robotics 2016, 32, 1309-1332, doi:10.1109/tro.2016.2624754.

[23] J. Sankaranarayanan, H. Samet and A. Varshney. A Fast k-Neighborhood Algorithm for Large Point-Clouds. In Proceedings of the Eurographics symposium on point-based graphics, Massachusetts, 2006; pp. 75-84.

[24] W. Zhang, J. Qi, P. Wan, H. Wang, D. Xie, X. Wang and G. Yan. An Easy-to-Use Airborne LiDAR Data Filtering Method Based on Cloth Simulation. Remote Sensing 2016, 8, doi:10.3390/rs8060501.

[25] A. Trevor, S. Gedikli, R. B. Rusu and H. I. Christensen. Efficient organized point cloud segmentation with connected components. Semantic Perception Mapping and Exploration (SPME) 2013, 10, 251-257.

[26] Y. Mo, Y. Wu, X. Yang, F. Liu and Y. Liao. Review the state-of-the-art technologies of semantic segmentation based on deep learning. Neurocomputing 2022, 493, 626-646, doi:10.1016/j.neucom.2022.01.005.

[27] T. Betsas, A. Georgopoulos, A. Doulamis and P. Grussenmeyer. Deep Learning on 3D Semantic Segmentation: A Detailed Review. Remote Sensing 2025, 17, doi:10.3390/rs17020298.

[28] S. Sarker, P. Sarker, G. Stone, R. Gorman, A. Tavakkoli, G. Bebis and J. Sattarvand. A comprehensive overview of deep learning techniques for 3D point cloud classification and semantic segmentation. Machine Vision and Applications 2024, 35, doi:10.1007/s00138-024-01543-1.

[29] Y. Wang, Y. Sun, Z. Liu, S. E. Sarma, M. M. Bronstein and J. M. Solomon. Dynamic Graph CNN for Learning on Point Clouds. ACM Transactions on Graphics 2019, 38, 1-12, doi:10.1145/3326362.

[30] E. Hameiri and I. Shimshoni. Estimating the principal curvatures and the Darboux frame from real 3D range data. In Proceedings of the Proceedings. First International Symposium on 3D Data Processing Visualization and Transmission, 2002; pp. 258-267.

[31] I. T. Jolliffe and J. Cadima. Principal component analysis: a review and recent developments. Philos Trans A Math Phys Eng Sci 2016, 374, 20150202, doi:10.1098/rsta.2015.0202.

[32] S. K. Singh, S. Raval and B. P. Banerjee. Automated structural discontinuity mapping in a rock face occluded by vegetation using mobile laser scanning. Engineering Geology 2021, 285, doi:10.1016/j.enggeo.2021.106040.

[33] W. Liu, J. Sun, W. Li, T. Hu and P. Wang. Deep Learning on Point Clouds and Its Application: A Survey. Sensors (Basel) 2019, 19, doi:10.3390/s19194188.

[34] B. Khemani, S. Patil, K. Kotecha and S. Tanwar. A review of graph neural networks: concepts, architectures, techniques, challenges, datasets, applications, and future directions. Journal of Big Data 2024, 11, doi:10.1186/s40537-023-00876-4.

[35] R. Q. Charles, H. Su, M. Kaichun and L. J. Guibas. PointNet: Deep Learning on Point Sets for 3D Classification and Segmentation. In Proceedings of the 2017 IEEE Conference on Computer Vision and Pattern Recognition (CVPR), 2017; pp. 77-85.

[36] C. R. Qi, L. Yi, H. Su and L. J. Guibas. PointNet++: deep hierarchical feature learning on point sets in a metric space. In Proceedings of the Proceedings of the 31st International Conference on Neural Information Processing Systems, Long Beach, California, USA, 2017; pp. 5105–5114.

[37] H. Thomas, C. R. Qi, J.-E. Deschaud, B. Marcotegui, F. Goulette and L. Guibas. KPConv: Flexible and Deformable Convolution for Point Clouds. In Proceedings of the 2019 IEEE/CVF International Conference on Computer Vision (ICCV), 2019; pp. 6410-6419.

[38] Q. Hu, B. Yang, L. Xie, S. Rosa, Y. Guo, Z. Wang, N. Trigoni and A. Markham. RandLA-Net: Efficient Semantic Segmentation of Large-Scale Point Clouds. In Proceedings of the 2020 IEEE/CVF Conference on Computer Vision and Pattern Recognition (CVPR), 2020; pp. 11105-11114.